\pdfoutput=1

\documentclass[11pt]{article}

\usepackage[]{ACL2023}

\usepackage{times}
\usepackage{latexsym}
\usepackage{booktabs}
\usepackage{multirow}
\usepackage{multicol}
\usepackage{enumitem}
\usepackage{floatrow}
\usepackage{subcaption}
\newfloatcommand{capbtabbox}{table}[][\FBwidth]
\usepackage[T1]{fontenc}

\usepackage[utf8]{inputenc}

\usepackage{microtype}

\usepackage{inconsolata}
\usepackage{bm}
\usepackage{amsmath}
\usepackage{graphicx}
\usepackage[normalem]{ulem}
\usepackage{hyperref}
\usepackage[hyphenbreaks]{breakurl}
\useunder{\uline}{\ul}{}
\newcommand{\muhao}[1]{{\color{blue}[{MC:} #1]}}
\newcommand{\resolve}[1]{#1}

\makeatletter
\def\UrlAlphabet{%
      \do\a\do\b\do\c\do\d\do\e\do\f\do\g\do\h\do\i\do\j%
      \do\k\do\l\do\m\do\n\do\o\do\p\do\q\do\r\do\s\do\t%
      \do\u\do\v\do\w\do\x\do\y\do\z\do\A\do\B\do\C\do\D%
      \do\E\do\F\do\G\do\H\do\I\do\J\do\K\do\L\do\M\do\N%
      \do\O\do\P\do\Q\do\R\do\S\do\T\do\U\do\V\do\W\do\X%
      \do\Y\do\Z}
\def\UrlDigits{\do\1\do\2\do\3\do\4\do\5\do\6\do\7\do\8\do\9\do\0}
\g@addto@macro{\UrlBreaks}{\UrlOrds}
\g@addto@macro{\UrlBreaks}{\UrlAlphabet}
\g@addto@macro{\UrlBreaks}{\UrlDigits}
\makeatother

\title{
Contrastive 
Bootstrapping for Label Refinement}

\author{
Shudi Hou$^{ \dagger}$ \and Yu Xia$^{\dagger}$ \and Muhao Chen$^{\ddagger}$ \and Sujian Li$^{\dagger}$\\
$^{\dagger}$Key Laboratory of Computational Linguistics, MOE, Peking University \\
$^{\ddagger}$University of Southern California\\ 
\texttt{\{housd, yuxia, lisujian\}@pku.edu};\; \texttt{{muhaoche@usc.edu}}\\
}

\begin{document}
\maketitle
\begin{abstract}

Traditional text classification typically categorizes 
texts into pre-defined coarse-grained classes,
from which the produced models cannot handle the real-world scenario where finer categories emerge periodically for accurate services.
In this work, we investigate the setting where fine-grained classification is done 
only using the
annotation of coarse-grained categories and the coarse-to-fine mapping.
We propose a lightweight contrastive clustering-based bootstrapping method to iteratively refine the labels of passages.
During clustering, it pulls away negative passage-prototype pairs under the guidance of the mapping from both global and local perspectives.
Experiments on NYT and 20News show that our method outperforms the state-of-the-art methods by a large margin.\footnote{Code is available at \url{https://github.com/recorderhou/contrastive\_bootstrapping\_label\_refinement}}
\end{abstract}

\section{Introduction}
\label{sec:intro}

Traditional text classification 
often categorize into a set of coarse-grained classes, which 
falls short in real-world scenarios where finer categories emerge. 
To this end, coarse-to-fine text classification is introduced \citep{mekala-etal-2021-coarse2fine}, which performs fine-grained classification given only annotation of coarse-grained categories and the coarse-to-fine mapping.
Then, 
it finetunes a pre-trained language model for each coarse prototype.\footnote{We use prototype and category interchangeably.}
However, this two-step method could be sub-optimal. 
For example, it is vulnerable to the noise which is propagated and accumulated through the pipeline.
Besides, it requires finetuning and saving a pre-trained language model for each coarse prototype which is heavyweight.

\begin{figure}[htbp]
    \centering
    \includegraphics[width=1.0\columnwidth]{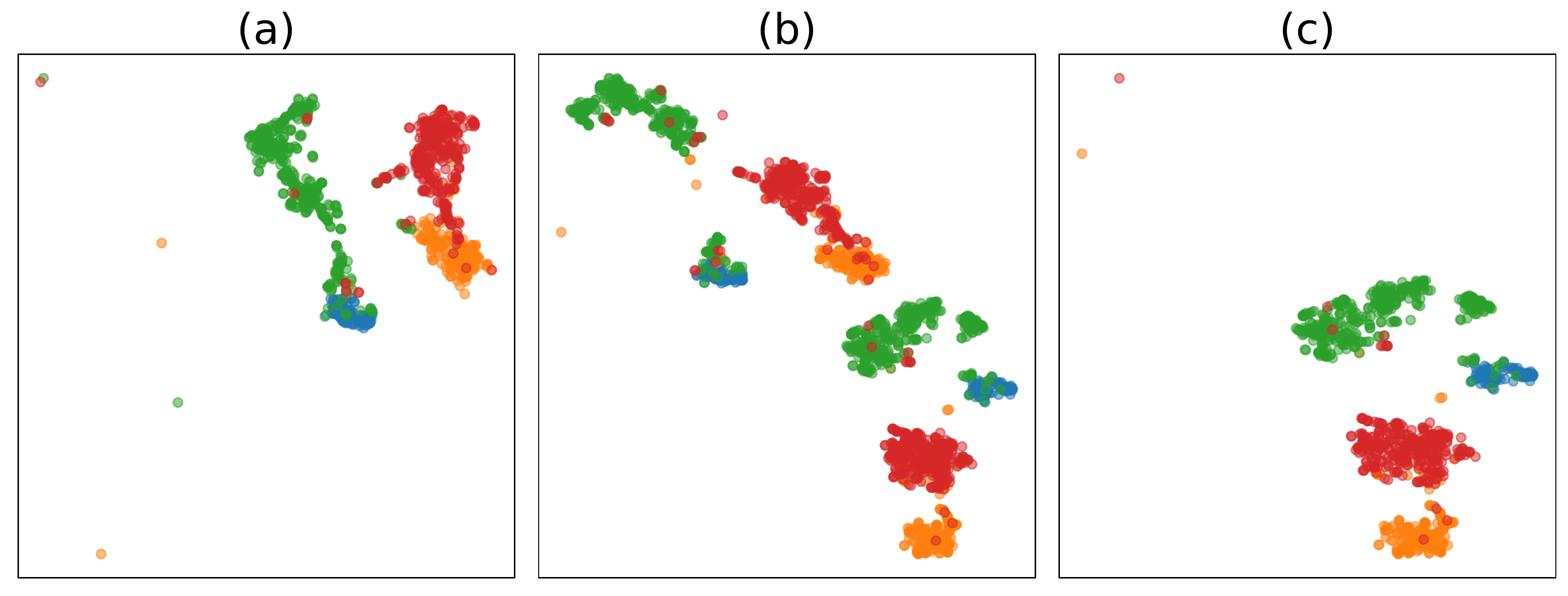}
    \caption{Passages with ``Arts'' coarse prototype on NYT dataset. Colors are used to denote different fine prototypes.
    (a) warm-up. (b) bootstrapping w/o selection strategy. (c) bootstrapping w/ selection strategy.}
    \label{fig:motivation}
\end{figure}

To this end, we propose a lightweight bootstrapping method based on contrastive clustering to iteratively refine the labels of passages.\footnote{We focus on passage-level classification as it is consistent with prior studies \cite{mekala-etal-2021-coarse2fine}. Though, without loss of generality, the studied problem as well as the proposed method can be extended to classifying natural language text in other granularities.}
To be more specific, the method starts with an epoch of warm-up on the weakly-labeled dataset.
During warm-up, it pulls away negative passage-prototype pairs under the guidance of the mapping from both global and local perspectives, \textit{i.e.}, coarse inter-cluster and fine inter-cluster perspectives.
After the warm-up, the distances between clusters are not significant which causes misclassification.
Instead of continuing training on the weakly-labeled dataset which might greatly increase the noise (Figure~\ref{fig:motivation}(b)),
we perform
a
bootstrapping 
process
which finetunes the model on the selected dataset and updates the selected dataset by the finetuned model alternately.
To mitigate the noise, we propose a selection strategy to identify high-quality pairs in terms of similarity and distinction.
To further boost our method, we adopt a modified similarity metric from \cite{lample2018word} and use the gloss knowledge to augment the prototype representation.
As shown in (Figure~\ref{fig:motivation}(c)), the resulting clusters are well separated with less noise.

Our contributions are summarized as follows: 
\begin{itemize}[noitemsep,topsep=0pt]
\item We propose a lightweight bootstrapping method based on contrastive clustering to address the problem of coarse-to-fine text classification.
\item Our method outperforms the state-of-the-art methods on two 
widely-used
datasets. Further analysis verifies the effectiveness of our proposed techniques.
\end{itemize}

\section{Proposed Method}

This section describes the technical details of the proposed method, starting with the task description.

\subsection{Task Description}
We follow the 
task definition of coarse-to-fine 
\resolve{text} classification in previous work \cite{mekala-etal-2021-coarse2fine}. 
Given $n$ passages $\{p_1, ..., p_n\}$ with their corresponding coarse-grained labels $\{c_1, ..., c_n\}$, along with the coarse-to-fine mapping $\mathcal{T}$, our goal is to assign a fine-grained label to each passage.
The key
notations used in our paper are defined as follows:
(1) $\mathcal{C}=\{\mathcal{C}_1, \mathcal{C}_2, ..., \mathcal{C}_m\}$ denotes the coarse prototypes.
(2) $\mathcal{F} = \{\mathcal{F}_1, \mathcal{F}_2, ..., \mathcal{F}_k\}$ denotes the fine prototypes.
(3) $\mathcal{T}: \mathcal{C} \to \mathcal{F}$ denotes the coarse-to-fine mapping, a 
surjective
mapping which separates $\mathcal{F}$ into $|\mathcal{C}|$ non-overlapping partitions.
(4) $\mathcal{S}_{pf}=\mathcal{T}(c_i)$ denotes the 
fine-grained candidate prototype
of $p_i$, 
which is also dubbed as $p$ for simplicity.
(5) $\mathcal{S}_{nf}= {\mathcal{F}} / {\mathcal{S}_{pf}}$ denotes fine prototypes not belonging to $\mathcal{T}(c_i)$.
(6) $\mathcal{S}_{nc}= {\mathcal{C}} / {c_i}$ denotes coarse prototypes in $\mathcal{C}$ other than $c_i$.


\subsection{Our Method}
\begin{figure}[t]
    \centering
    \includegraphics[width=0.8\textwidth]{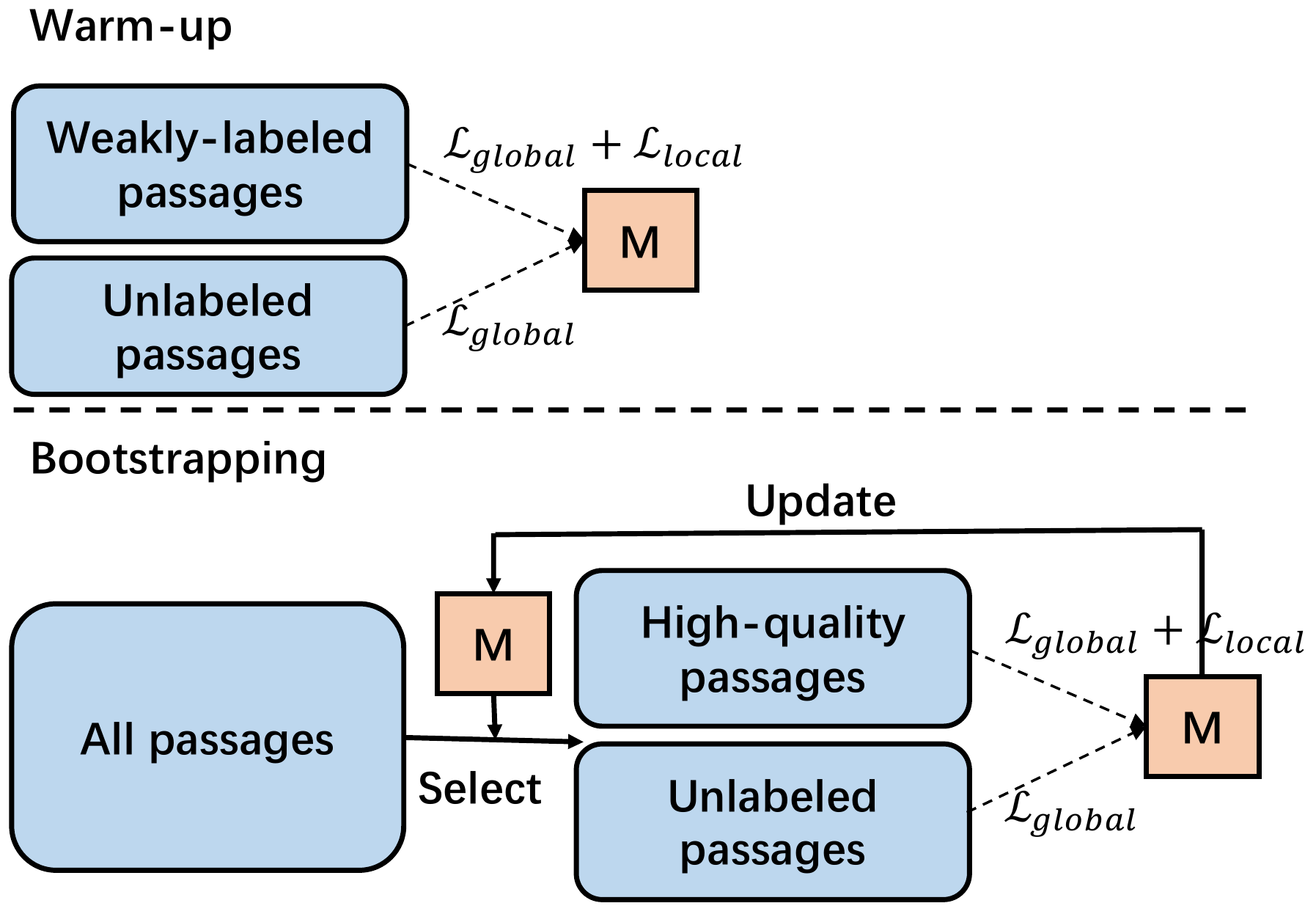}
    \caption{Illustration of our training process.}
    \label{fig:overview}
\end{figure}

\paragraph{Training Process} 
As illustrated in Figure~\ref{fig:overview}, we start with an epoch of warm-up, during which we optimize 
two contrastive losses
$\mathcal{L}_{global}$, $\mathcal{L}_{local}$ on the weakly-labeled dataset and 
only the
$\mathcal{L}_{global}$ on the unlabeled dataset.
The two contrastive losses are detailed in the following paragraphs.
Then, we 
conduct
several epochs of bootstrapping with the above model.
At each bootstrapping step, we first select 
a small set of passages on which labels are predicted with high confidence
by the model.
Then, we finetune the model on the selected dataset with the same losses as 
warm-up.
We repeat the finetuning and the selection alternately.

\paragraph{Initial Weak Supervision}
Following previous work, we consider samples that exclusively contain the
label surface name as their respective weak supervision. 
More details can be referred to the 
prior study.

\paragraph{Passage and Prototype Representation}
We encode passages $\{p_1, ..., p_n\}$ and all prototypes $\mathcal{C}\cup\mathcal{F}$ into the same 
embedding
space \resolve{with a pretrained language model}.
The resulting passage representation and prototype representation are denoted as $\bm{p}$ and $\bm{l}$ respectively.
During the training process, the prototype representations are dynamically updated
to fit the current passage representations.
Specifically, we use the last hidden representation of 
\textsc{[CLS]} as their representations.

\paragraph{Similarity Metric}
Cosine similarity is often used to measure semantic 
similarity of embedding representations.
However, in high-dimensional spaces, some ``hub'' vectors may be close to many other vectors while 
some
other vectors are 
instead being
isolated.
For example, a passage's representation $\bm{p}$ may get high cosine with 
a large number of labels in $S_{pf}$ 
due to such hubness issues.
In this case, a high similarity score does not necessarily 
lead to
a high discrepancy among labels.
Selecting a highly-scored label from the hub as the seed is potentially detrimental to our pairing-based method.
Inspired by cross-domain similarity local scaling \citep{lample2018word}, we adopt a modified similarity metric $c(\bm{p}, \bm{l})$ to prevent passage vectors from becoming hubs:
\begin{equation}
\begin{aligned}
c(\bm{p}, \bm{l}) = \cos(\bm{p}, \bm{l}) - KNN(\bm{p})
\end{aligned}
\label{1.1}
\end{equation}
\vspace{-1em}
\begin{equation}
\begin{aligned}
KNN(\bm{p}) = \frac{1}{K}\sum\max_{\bm{l} \in \mathcal{F}}K\{\cos(\bm{p}, \bm{l})\}
\end{aligned}
\label{1.2}
\end{equation}
where 
$KNN(.)$
denotes $K$ nearest neighbors.

\begin{figure}
    \centering
    \includegraphics[width=0.6\columnwidth]{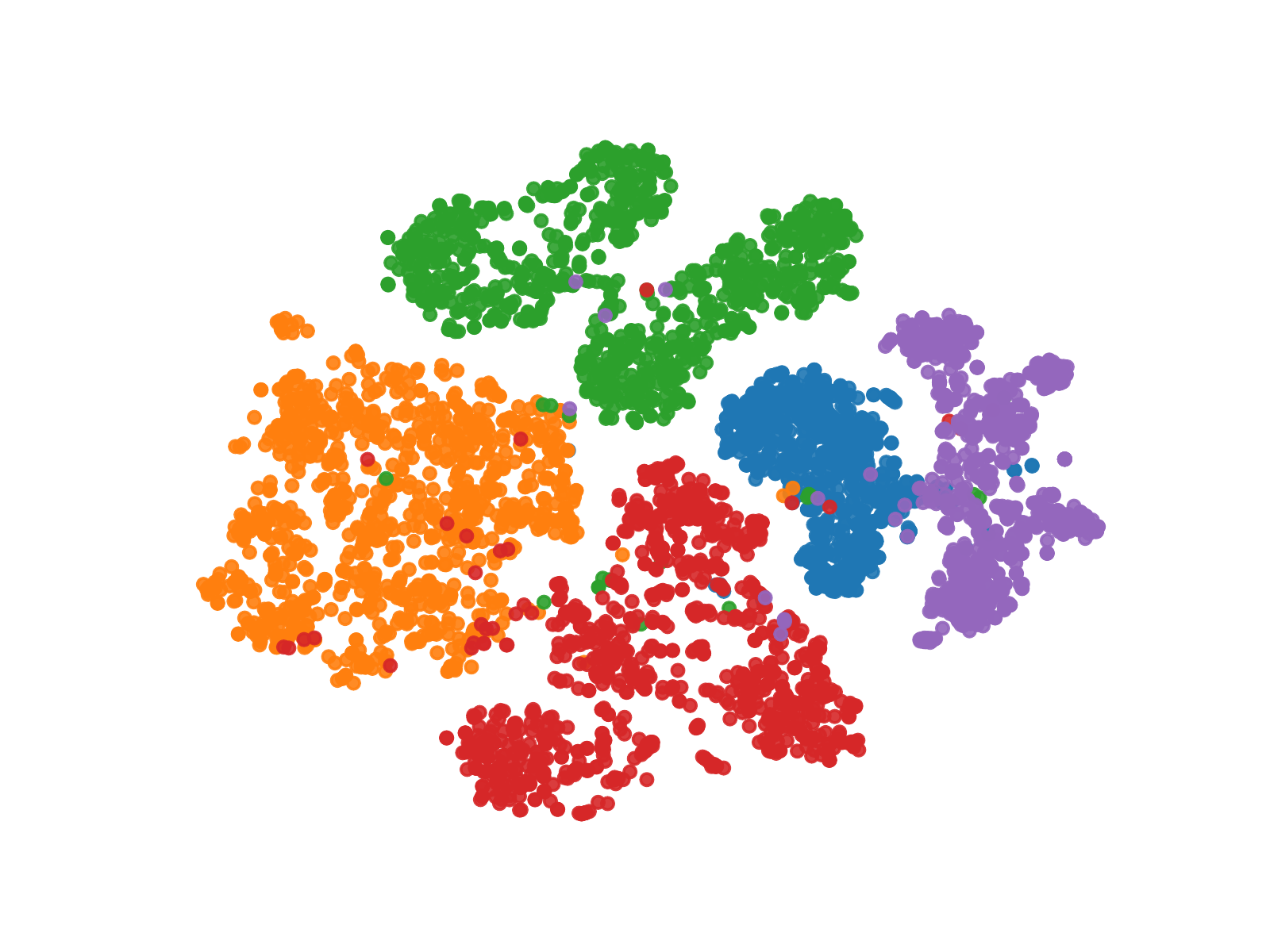}
     \caption{Passage representations after warm-up on 20News dataset. Colors are used to denote different coarse prototypes.}
    \label{fig:20news_coarse}
\end{figure}

\paragraph{Warm-up}
Viewing a passage as an anchor, we 
expect
that its semantic similarity to the 
correct fine-grained prototype
should be closer than any other 
fine-grained candidate
prototypes.
We regard the distance in the representation space as the similarity.
Specifically, we optimize the following margin ranking loss:
\begin{equation}
\small
\begin{aligned}
\mathcal{L}_{global} = \frac{1}{|S_{pf}|}\sum_{\substack{l\in S_{pf}\\  l'\in S_{nf}}} \max\{c(\bm{p}, \bm{l}) - c(\bm{p}, \bm{l'}) + \gamma, 0\}
\end{aligned}
\label{warmup}
\end{equation}
where $\gamma$ is a hyper-parameter denoting the margin.
We use all fine candidate prototypes in $S_{pf}$ as positive examples and randomly sample the same number of prototypes from $S_{nf}$ as negative examples.
We view this loss as a global loss to cluster samples according to their coarse labels (Figure~\ref{fig:20news_coarse}).


For instances labeled in the initial weak supervision stage, we adopt another margin ranking loss: 
\begin{equation}
\begin{aligned}
\mathcal{L}_{local} = \max\{sec\_max - c(\bm{p}, \bm{l}) + \sigma, 0\}
\end{aligned}
\label{confidence}
\end{equation}
\vspace{-1em}
\begin{equation}
\begin{aligned}
sec\_max = \max_{l' \in S_{pf}, l' != l}c(\bm{p}, \bm{l'})
\end{aligned}
\label{second_max}
\end{equation}
We 
regard
this loss as a local loss to cluster samples according to their 
fine-grained
labels (Figure~\ref{fig:motivation} (a)).

\paragraph{Bootstrapping}
After the warm-up, representations show an inclination to form clusters. \resolve{Yet,} the distances between them are not significant 
\resolve{enough to separate the classes}.
To further get compact clusters, we perform bootstrapping which finetunes the model on the selected dataset and updates the selected dataset by the finetuned model alternately.
Instead of using the initial weak supervision which might greatly increase the noise as observed,
we propose a selection strategy to select high-quality passage-prototype pairs.
Specifically, we assign a pseudo label to each passage by their similarity (Eq.\eqref{sim}).
Apart from \textbf{similarity}, we assume high-quality pairs should also be \textbf{discriminative} (Eq.\eqref{uniq}):
\begin{equation}
\begin{aligned}
l = arg\max_{l \in S_{pf}} c(\bm{p}, \bm{l})
\end{aligned}
\label{sim}
\end{equation}
\begin{equation}
\begin{aligned}
c(\bm{p}, \bm{l}) - \max_{l'\in S_{pf}, l' != l}c(\bm{p}, \bm{{l'}}) > \beta
\end{aligned}
\label{uniq}
\end{equation}
where $\beta$ is a threshold updated at each epoch.
We construct a confident set $\mathrm{CS}$ with top $r\%$ pairs satisfying these two conditions.
We update $\beta$ with the lowest similarity in $\mathrm{CS}$.
Then, we optimize Eq.\eqref{confidence} and Eq.\eqref{warmup} on $\mathrm{CS}$ and the rest passages accordingly.
\paragraph{Gloss Knowledge}
Since the surface names alone can not well represent the semantics of labels, we enrich them with external semantic knowledge.
To be more specific, we select the first two sentences in each surface name's first Wikipedia webpage to augment the original surface name with a predefined template (Table~\ref{table:template}).
We adopt the format of ``template, surface name, gloss'' and use the last hidden representation of [CLS] as their representation.
\paragraph{Prediction} It is worth noticing that applying our similarity metric $c(\bm{p}, \bm{{l}})$ do not change the relative ranking among labels in $\mathcal{S}_{pf}$ compared with the cosine similarity. For simplicity, we use cosine similarity for prediction. 
\begin{equation}
\begin{aligned}
l = arg\max_{l \in S_{pf}} \cos(\bm{p}, \bm{l})
\end{aligned}
\label{prediction}
\end{equation}

\section{Experiments}

In this section, we describe the experimental evaluation for the proposed method.

\subsection{Datasets and Metrics}
For a fair comparison with prior work, we use the same hierarchical datasets used by
We report both Macro-F1 and Micro-F1 for evaluation on the following two datasets.
\paragraph{The 20 Newsgroups (20News)} The passages in 20News was organized into 5 coarse-grained newsgroups and 20 fine-grained newsgroups corresponding to different topics (Table~\ref{tab:statistic}). Passages in 20News were partitioned evenly across the 20 different fine-grained newsgroups.\footnote{\url{http://qwone.com/~jason/20Newsgroups/}} Following \cite{mekala-etal-2021-coarse2fine}, we omitted the 3 miscellaneous newsgroups (``misc.forsale,'' ``talk.politics.misc'' and ``talk.religion.misc'') and expanded the abbreviation to full words. 
\paragraph{The New York Times (NYT)} 
This dataset contains 5 coarse-grained topics and 25 subtopics (Table~\ref{tab:statistic}). The NYT dataset is highly skewed with the coarse-grained topic ``sports'' containing more than 80\% passages.



\begin{table*}[]
\centering
\small
\begin{tabular}{l|ll|ll}
\toprule
              & \multicolumn{2}{c|}{NYT}        & \multicolumn{2}{c}{20News}      \\
              & Mi-F1(\%)      & Ma-F1(\%)      & Mi-F1(\%)      & Ma-F1(\%)      \\ 
              \midrule
LOT-Class     &     79.26           &    63.16            &      56.38          &      54.80          \\
X-Class       & 58.15          & 60.50          & 52.95          & 53.47          \\
C2F           & 89.23          & 84.36          & 75.77          & 75.24          \\ 
C2F w/ our select $\star$         &  89.64         & 82.72         &  77.20         & 76.41          \\
\midrule
Ours          & \textbf{92.64} & \textbf{89.90} & {\ul 77.64}    & {\ul 77.22}    \\
w/o fine & 91.15 ($\downarrow$ 1.49)    &  84.90 ($\downarrow$ 5.00)    & 74.34 ($\downarrow$ 3.30)         & 73.78 ($\downarrow$ 3.44)          \\
w/o bootstrap & 89.49 ($\downarrow$ 3.15)         & 82.50 ($\downarrow$ 7.40)         & 76.01 ($\downarrow$ 1.63)         & 75.46 ($\downarrow$ 3.30)         \\
w/o gloss     & 89.91 ($\downarrow$ 2.73)         & 80.48  ($\downarrow$ 9.42)        & 72.68 ($\downarrow$ 4.86)          & 70.31 ($\downarrow$ 6.91)       \\
w/o select    & 87.56 ($\downarrow$ 5.08)          & 81.98  ($\downarrow$ 8.02)        & \textbf{79.74} ($\uparrow$ 2.10) & \textbf{79.21} ($\uparrow$ 1.99) \\
w/o similarity     & 89.25 ($\downarrow$ 3.39)         & 82.44 ($\downarrow$ 7.46)         & 61.21 ($\downarrow$ 16.43)         & 54.76 ($\downarrow$ 22.46)         \\ 
w/ Manhattan similarity $\dagger$    & 33.45 ($\downarrow$ 59.19)         & 39.47 ($\downarrow$ 50.43)         & 41.83 ($\downarrow$ 35.81)         & 36.50 ($\downarrow$ 40.72)         \\ 
w/ Euclidean similarity $\ddagger$    & 
{\ul 92.46} ($\downarrow$ 0.18)         & 
{\ul 89.17} ($\downarrow$ 0.73)         & 72.11 ($\downarrow$ 5.53)         & 70.65 ($\downarrow$ 6.57)         \\
\bottomrule
\end{tabular}%
\caption{Results on NYT and 20News. ``$\star$" equips C2F with our selection strategy. ``$\dagger$'' replaces our similarity metric with Manhattan distance. ``$\ddagger$'' replaces our similarity metric with Euclidean distance.}
\label{baselines}
\end{table*}

\subsection{Main Results}
We compare our model with the previous work \citep{mekala-etal-2021-coarse2fine}, as well as 
several
zero-shot weakly supervised text classification methods \citep{wang-etal-2021-x, meng-etal-2020-text} following previous works.
We reproduce them using their implementation.\footnote{\url{https://github.com/yumeng5/LOTClass}}\footnote{\url{https://github.com/ZihanWangKi/XClass}}\footnote{\url{https://github.com/dheeraj7596/C2F}}

As shown in Table~\ref{baselines}, our method outperforms the baselines by 5.67\% in Micro-F1 and 5.54\% in Macro-F1 on the NYT dataset, as well as 3.97\% in Micro-F1 and 3.04\% in Macro-F1 on 20News dataset.

\subsection{Analysis}
To verify the effectiveness of 
different model components
, we 
conduct ablation studies to test each of those.

\paragraph{Effect of Bootstrapping}
The ``w/o bootstrap'' results in
Table~\ref{baselines} report the performance with warm-up only.
These
results are consistently lower than those with bootstrapping.
Specifically,
bootstrapping improves the warm-up by 3.15\% Micro-F1, 7.40\% Macro-F1 and 1.63\% Micro-F1, 3.30\% Macro-F1 on NYT and 20News respectively.
Figure~\ref{fig:motivation}(a)(c) shows passage representations are more separated from each other.

\paragraph{Effect of Selection Strategy}
We replace the selection strategy in bootstrapping with the initial weakly-labeled samples.
From the 
``w/o bootstrap''
results in Table~\ref{baselines}, 
we can see that, our selection strategy brings an improvement of 4.26\% Micro-F1, 7.46\% Macro-F1 on NYT.
It is better to use the seed dataset on 20News.
We hypothesize 
that this observation
is because the seed dataset has a more balanced label distribution than our selected high-quality samples on 20News. 
We also incorporate our selection strategy to the C2F baseline in the bootstrapping stage. As shown in Table~\ref{baselines} row ``C2F w/ our select,'' this strategy improves the performance of C2F by 1.43\% Micro-F1, 1.17\% Macro-F1 on 20News and 0.41\% Micro-F1 on NYT, exhibiting the effectiveness of  our strategy.

\begin{figure*}[t]
    \centering
    \includegraphics[width=0.84\textwidth]{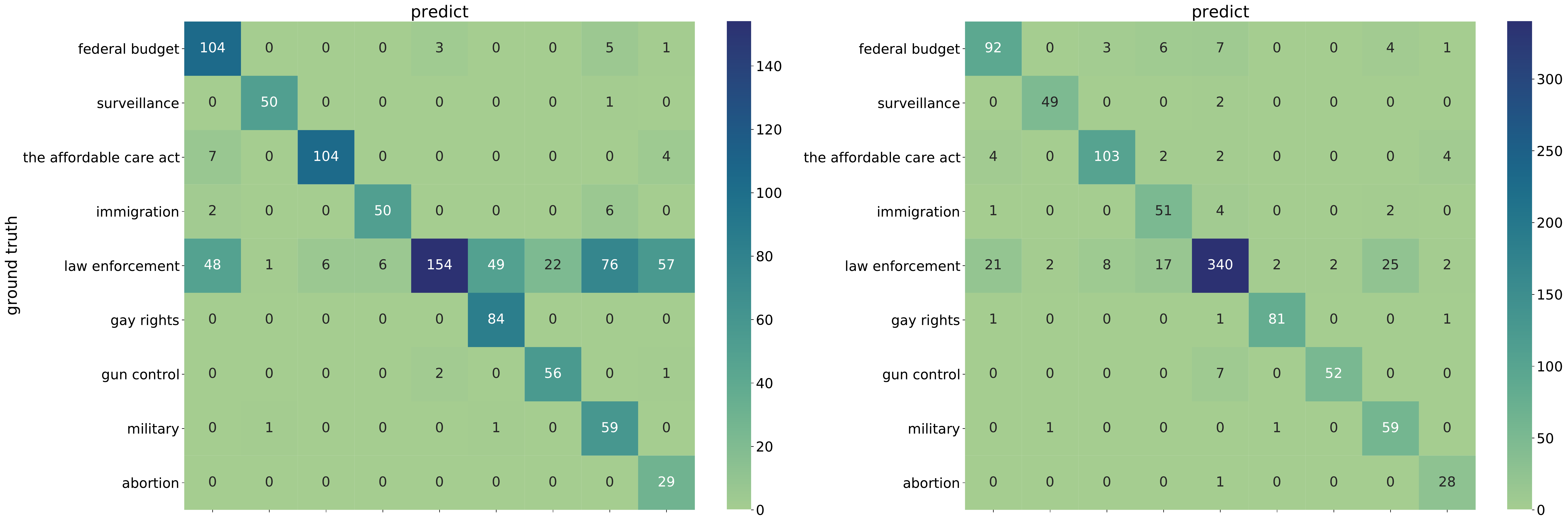}
    \caption{Confusion matrix on ``Politics'' coarse prototype. 
    Our similarity metric (right) outperforms cosine similarity (left) by 12.25\% Macro-F1 and 16.68\% Micro-F1 under ``Politics.''
    }
    \label{fig:csls_effect}
\end{figure*}

\paragraph{Effect of Similarity Metric} 
We replace our similarity metric with the cosine similarity.
From Table~\ref{baselines} ``w/o similarity'' we can see that, our similarity metric brings along an improvement of 3.39\% in Micro-F1, 7.46\% in Macro-F1 on NYT, and 16.43\% in Micro-F1 and 22.46\% in Macro-F1 on 20News.
From Figure~\ref{fig:csls_effect},  
we can see that 63\% of samples belonging to the ``Law Enforcement'' prototype are misclassified using the cosine similarity.
However, 18\% are misclassified using our similarity metric, verifying its effectiveness.
Besides, results for
``w/ Manhattan similarity''
and ``w/ Euclidean similarity'' show that alternating cosine similarity in $c(\bm{p}, \bm{l})$ causes performance drops of 35.81\% (5.53\%) in Micro-F1, 40.72\% (6.57\%) in Macro-F1 and 50.19\% (0.18\%) in Micro-F1, 50.43\% (0.73\%) in Macro-F1 on 20News and NYT data, further proving the effectiveness of our similarity metric. 

\paragraph{Effect of Gloss Knowledge} 
We remove the gloss knowledge and use the label surface name only.
Comparing the ``w/o gloss'' results in Table~\ref{baselines} with the full-setting ones, 
we observe that the gloss knowledge brings an improvement of 2.73\% in Micro-F1, 9.42\% in Macro-F1 on NYT and 4.86\% in Micro-F1, 6.91\% in Macro-F1 on 20News.
Figure~\ref{fig:gloss_effect} further shows 
the effect of gloss knowledge
on different prototypes.


\paragraph{Extending to the setting without coarse-to-fine mapping}
We extend our method to the setting without the coarse-to-fine mapping.
In other words, the only supervision is the gold coarse labels.
We modify $\mathcal{L}_{global}$ as follows: 
\begin{equation}
\begin{aligned}
\mathcal{L}_{c\_global} = \max\{c(\bm{p}, \bm{l_c}) - c(\bm{p}, \bm{l_c'}) + \gamma, 0\}
\end{aligned}
\label{c_warmup}
\end{equation}
where we use the golden coarse label $\bm{l_c}$ as the positive example and randomly sample one coarse label $\bm{l_c'}$ from $\mathcal{S}_{nc}$ as the negative example. 
The ``w/o fine'' results in Table~\ref{baselines} show that the performance does not degrade much when the association between coarse and fine-grained labels does not exist,
showing the feasibility of our method in a more general setting.

\section{Related Work}

Previous works in weakly supervised text classification have explored different kinds of weak supervision.
(1) a set of related keywords.
\cite{mekala2020contextualized} augment and disambiguate the initial seed words with contextualized and highly label-indicative keywords.
\cite{meng2020text} identify keywords for classes by querying replacements for class names using
BERT and pseudo-labels the documents by heuristics with the selected keywords.
(2) a few labeled documents.
\cite{tang2015pte} represent the labeled documents and different
levels of word co-occurrence information as a large-scale text network.
 \cite{meng2018weakly} propose a pseudo-document generator that leverages the seed labeld documents to generate pseudo-labeled documents for model pre-training.
(3) label surface names.
\cite{wang-etal-2021-x} propose an adaptive representation learning method to obtain label and document embedding, and cluster them to pseudo-label the corpus.
Our setting is different from theirs in that we use coarse-grained annotation to improve the fine-grained text classification. 

Contrastive learning \citep{He_2020_CVPR, pmlr-v119-chen20j, NEURIPS2020_d89a66c7} aims at learning representations by contrasting the positive pairs and negative pairs. In NLP, existing works can be primarily categorized into two distinct streams.
Unsupervised contrastive learning seeks to contrast grouped or perturbed instances to generate more robust representation of unlabeled textual data \citep{gao-etal-2021-simcse, wei2021on, kim-etal-2021-self, wang-etal-2021-cline}.
On the contrary, supervised contrastive learning \citep{suresh-ong-2021-negatives, zhou-etal-2021-contrastive, yu-etal-2021-fine, huang-etal-2022-unified} is label-aware 
and seeks to create representations for differently labeled data with more discrepancy.
Our work has shown that supervised contrastive learning incorporating label names, with minimal external knowledge, improves the model’s performance in label refinement. 

\section{Conclusion}
In this paper, we study the task of coarse-to-fine text classification.
We propose a novel contrastive clustering-based bootstrapping method to refine the label in an iterative manner.
Experiments on two real-world datasets 
for coarse-to-fine text classification
verify the effectiveness of our method.
Future work could consider extending this method to other fine-grained decision-making tasks that could potentially benefit from coarse-grained labels, such as various kinds of lexical semantic typing tasks \cite{huang-etal-2022-unified}.
Another meaningful direction is to consider incorporating other partial-label learning techniques \cite{10.1145/2939672.2939788} that are relevant to coarse-to-fine prediction tasks.

\section*{Limitations}
Our paper has the following limitations:
(1) In real-world applications, the label hierarchy may be more than two levels.
It is worth extending our method to such a setting and empirically verifying it.
(2) Our selection strategy simply takes top $r\%$ confident samples, which might result in class imbalance problem. 
Alleviating the imbalance problem may further improve our performance.
We leave them as future work.

\section*{Acknowledgement}

We appreciate the reviewers for their insightful
comments and suggestions.
We would like to express our gratitude to the authors of the C2F paper \cite{mekala-etal-2021-coarse2fine} for their collective effort in open-sourcing the dataset and code. 
Their released materials played a vital role in our research.

Shudi Hou, Yu Xia and Sujian Li were supported by  
National Key R\&D Program of China (No. 2020AAA0109703).
Muhao Chen was supported by the National Science Foundation of United States Grant IIS 2105329, a subaward of the INFER Program through UMD ARLIS, an Amazon Research Award and a Cisco Research Award.

\bibliography{anthology,custom}
\bibliographystyle{acl_natbib}

\appendix
\section{Dataset Statistics}
We list the statistics of the datasets in Table~\ref{tab:statistic}.
\label{sec:statistic}
\begin{table*}[]
\centering
\resizebox{\textwidth}{!}{%
\begin{tabular}{cccccc}
\toprule
Dataset & Passage & $|\mathcal{C}|$ & $|\mathcal{F}|$ & Coarse Prototype & Fine Prototype \\ \midrule
20News & 16468 & 5 & 17 & computer, politics, recreation, & graphics, windows, ibm, mac, x window, mideast, guns, autos, motorcycles, \\
 &  &  &  & religion, science & baseball, hockey, christian, atheism, encryption, electronics, medicine, space \\ \hline
 &  &  &  &  & dance, music, movies, television, economy, energy companies, international \\ 
NYT & 11744 & 5 & 26 & arts, business, politics, & business, stocks and bonds, abortion, federal budget, gay rights, gun control, \\
 &  &  &  & science, sports & immigration, law enforcement, military, surveillance, the affordable care act, \\
 &  &  &  &  & cosmos, environment, baseball, basketball, football, golf, hockey, soccer, tennis \\ \bottomrule
\end{tabular}%
}
\caption{Dataset Statistics.}
\label{tab:statistic}
\end{table*}

\section{Templates}
We list the templates used in Table~\ref{table:template}.

\begin{table*}
    \centering
    \small
    \resizebox{\textwidth}{!}
    {
    \begin{tabular}{ll}
    \toprule 
    Dataset & Template\\
	\midrule
         NYT & {1 : The news is about, 2 : The news is related to, 3 : The topic of this passage is}\\
        \midrule
        20News & {1 : The topic of this post is , 2 : They are discussing , 3 : This post mainly talks about}\\
	\bottomrule
    \end{tabular}}
    \caption{Three variants of templates used to concatenate the gloss knowledge and the surface name.
    The first template is best for NYT and the third template is best for 20News.}
    \label{table:template}
\end{table*}

\section{Effect of gloss knowledge on different prototypes}
\begin{figure*}[htbp]
    \centering
    \includegraphics[width=1.0\textwidth]{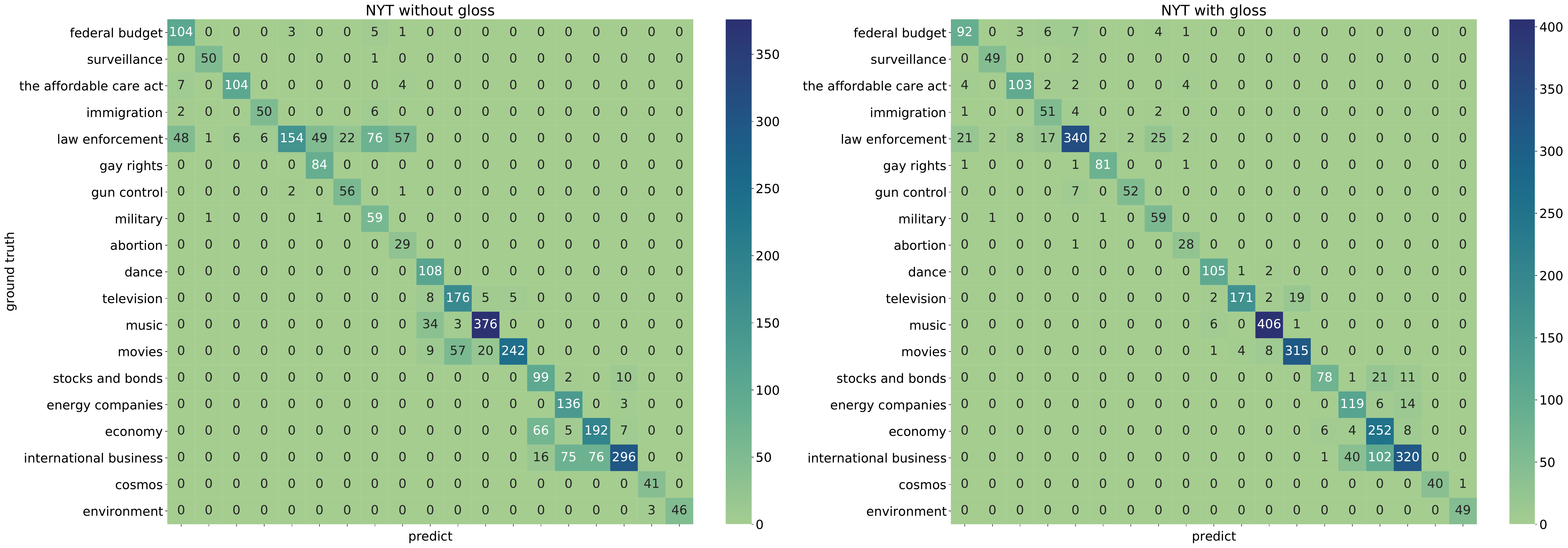}
    \caption{Confusion matrix over all fine prototypes without (left) and with (right) the gloss knowledge.}
    \label{fig:gloss_effect}
\end{figure*}

We show the confusion matrix over all fine prototypes in Figure~\ref{fig:gloss_effect}.

\section{Implementation Details}
We use RoBERETa-base \cite{Liu2019RoBERTaAR} as the encoder. 
The models are trained on one GeForce RTX 3090 GPU.
We set the batch size as 8.
We do one epoch of warmup and four epochs of bootstrapping. 
We use the predictions from the last epoch as the final predictions. 
We use AdamW \cite{Loshchilov2017FixingWD} as the optimizer.
$r$ is set as 15 for NYT and 1 for 20News. $\gamma$ and $\sigma$ are set as 0.05 for both NYT and 20News. We run our model 3 times using different random seeds. We used t-SNE \cite{scikit-learn, sklearn_api} for the visualization in this paper.

\section{Selection of $r$}
We select the value of $r$ from set \{1, 5, 10, 15, 20\}. For each coarse prototype $\mathcal{C}_i$, we calculate the ratio of initial weak supervision $W_{\mathcal{C}_i}$ in category $\mathcal{C}_i$ to the total number of instance $I_{\mathcal{C}_i}$ in $\mathcal{C}_i$, we denote the ratio as $R_{\mathcal{C}_i} = W_{\mathcal{C}_i} / I_{\mathcal{C}_i}$. After that, we select the $r$ closest to 
$\underset{\mathcal{C}_i \in \mathcal{C}}{\min}\{R_{\mathcal{C}_i}\}$. As shown in Table~\ref{tab:nyt-weaksup} and Table~\ref{tab:20news-weaksup}, the minimal $R_{\mathcal{C}_i}$ in NYT dataset is 13.43\%, closest to 15, while the minimal $R_{\mathcal{C}_i}$ in 20News dataset is 2.05\%, closest to 1.

\begin{table*}[htbp]
  \centering
  \begin{subtable}{0.45\textwidth}
    \centering
    \begin{tabular}{cccc}
    \toprule
    $\mathcal{C}_i$ & $W_{\mathcal{C}_i}$ & $I_{\mathcal{C}_i}$ & $R_{\mathcal{C}_i}$ (\%)\\
    \midrule
    arts & 184 & 1043 & 17.64 \\ \midrule
    business & 132 & 983 & 13.43 \\ \midrule
    politics & 216 & 989 & 21.84 \\ \midrule
    science & 42 & 90 & 46.67 \\ \midrule
    sports & 1890 & 8639 & 21.88 \\
    \bottomrule
    \end{tabular}
    \caption{Ratio of the initial weak supervision in NYT}
    \label{tab:nyt-weaksup}
  \end{subtable}
  \hfill
  \begin{subtable}{0.45\textwidth}
    \centering
   \begin{tabular}{ccccc}
    \toprule
    $\mathcal{C}_i$ & $W_{\mathcal{C}_i}$ & $I_{\mathcal{C}_i}$ & $R_{\mathcal{C}_i}$ (\%)\\
    \midrule
    computer & 100 & 4880 & 2.05 \\ \midrule
    politics & 56 & 1850 & 3.03 \\ \midrule
    recreation & 924 & 3976 & 23.24 \\ \midrule
    religion & 150 & 1976 & 8.35 \\ \midrule
    science & 100 & 3951 & 2.53 \\
    \bottomrule
    \end{tabular}
    \caption{Ratio of the initial weak supervision in 20News}
    \label{tab:20news-weaksup}
  \end{subtable}
   \caption{Ratio of the initial weak supervision}
\end{table*}

\end{document}